\documentclass[runningheads]{llncs}
\usepackage[colorlinks=true, urlcolor=blue, linkcolor=red]{hyperref}
\usepackage{multirow}
\usepackage{color}
\usepackage{cite}
\usepackage{graphicx}
\usepackage{listings}
\usepackage{algorithm}
\usepackage{algorithmic}
\usepackage{subfigure}
\usepackage{booktabs}
\usepackage{tikz}
\usepackage{makecell}
\usepackage{amssymb}
\def\checkmark{\tikz\fill[scale=0.4](0,.35) -- (.25,0) -- (1,.7) -- (.25,.15) -- cycle;} 

\DeclareUnicodeCharacter{2212}{-}
\begin{document}
\title{Delving into Ipsilateral Mammogram Assessment under Multi-View Network}
%------------AUTHOR INFORMATION AND AFFILIATIONs-----------------------------
\author{Toan T. N. Truong\inst{1\star} \and
Huy T. Nguyen\inst{2}\thanks{Toan T. N. Truong and Huy T. Nguyen equally contribute to this paper.} \and
Thinh B. Lam\inst{3} \and Duy V. M. Nguyen\inst{4} \and Phuc H. Nguyen\inst{5}}
% %
% \authorrunning{Huy T. Nguyen et al.}
\institute{
Ho Chi Minh City International University, Vietnam.
\and
National Cheng Kung University, Taiwan
\and
Ho Chi Minh City University of Science, Vietnam.
\and
Military Hospital 175, Vietnam
\and
Eastern International University, Vietnam
 \\
\email{\{phuc.nguyenhong@eiu.edu.vn\}}}
%---------------------END AUTHOR--------------------
% \author{Anonymous}
%
\authorrunning{Toan T. N. Truong et al.}
% \institute{Anonymous}

%
\maketitle              % typeset the header of the contribution
\begin{abstract}
%This paper investigates the impact of breast density distribution on the generalization performance of deep-learning models on mammography images using the VinDr-Mammo dataset. We explore various strategies when fusing them in different ways: average and concatenate. Furthermore, we also concern about how the model learns in different individuals and fusion pathways which assigns more or fewer layers in a Coarse Layer (low-level individual extractor ) and Fine Layer (high-level fusion extractor). Our Ipsilateral Multi-View Network contains total of five fusion types which are separated into several positions in ResNet18: Pre Fusion, Early Fusion, Middle Fusion, Last Fusion, and Post Fusion. Our results show that the Middle Fusion in concatenate fusion block, the most balanced fusion type, achieves the best method which improves the generalization performance of deep-learning models (+5.9\%) on concatenate and (+6.63\%) on average. This paper highlights the importance of how we assign layers in architecture to extract the multi-view network, particularly in the context of breast density distribution, which is critical in mammography screening.

In many recent years, multi-view mammogram analysis has been focused widely on AI-based cancer assessment. In this work, we aim to explore diverse fusion strategies (average and concatenate) and examine the model's learning behavior with varying individuals and fusion pathways, involving Coarse Layer and Fine Layer. The Ipsilateral Multi-View Network, comprising five fusion types (Pre, Early, Middle, Last, and Post Fusion) in ResNet-18, is employed. Notably, the Middle Fusion emerges as the most balanced and effective approach, enhancing deep-learning models' generalization performance by +2.06\% (concatenate) and +5.29\% (average) in VinDr-Mammo dataset and +2.03\% (concatenate) and +3\% (average) in CMMD dataset on macro F1-Score. The paper emphasizes the crucial role of layer assignment in multi-view network extraction with various strategies.\
% Our code is available here: \href{github.com/annonymous0811/Annonymous}{github.com/annonymous0811/Annonymous}
\end{abstract}
\section{Introduction}

%In 2020, breast cancer has beaten lung cancer to become the most commonly diagnosed cancer in women. The biggest specialized agency of the United Nations responsible for international public health, the World Health Organization (WHO), also states that claim, according to data published by the International Agency for Research on Cancer (IARC) in December 2020 \cite{who}. Therefore, one of the effective methods for alleviating the spread out of the disease is the early detect the masses in the patient's breast.

% , breast cancer has surpassed lung cancer to become the most frequently diagnosed cancer. In 2020, it is estimated that 2,3 million women will be diagnosed with breast cancer and 685 thousand will die from the disease. Late-stage breast cancer is frequently characterized by the onset of symptoms, but treatments may encounter difficulties during this period. Therefore, routine breast cancer screening is essential for the detection of breast tumors at an early stage. Mammography, a form of breast X-ray examination, is utilized in computer-aided diagnosis (CADx) systems to increase the efficacy of radiologists.
In recent years, many machine learning methods based on texture descriptors or deep learning networks have been proposed for classification using ipsilateral views. A mammographic screening normally consists of two views: craniocaudal view (CC), a top-down view of the breast, and mediolateral oblique (MLO), a side view of the breast taken at a certain angle. Radiologists examine the two views from both patient's breasts left and right, which are both views of the same breast (ipsilateral views) and the same view of both breasts (bilateral views). \emph{Y. Chen et.al.} \cite{local-global} proposed two pathways to extract the global feature and local feature between two ipsilateral views (CC and MLO). This methodology achieved good and desirable results. Continuously, \emph{Liu et.al.} \cite{actlikeradiologist1, actlikeradiologist2} successfully applied the well-known graph convolutional network (GCN) to the mammographic field which processes the bipartite GCN and inception GCN individually. Then, they fused both of them together in a correspondence reasoning enhancement stage to procedure the prediction. There are many other multi-view-based approaches\cite{anovel, multiviewlocalglobal, multiview1, multiview2, multiview3} using from two to four images as inputs.

Recent research has demonstrated that multi-perspective approaches \cite{wu, khan} and \cite{Geras} enhance breast cancer diagnosis. The main technique underpinning these approaches is the development of an end-to-end deep learning classification model for mammographic pathology. Before combining four screening mammogram views for prediction, this strategy extracts features from each view separately. Also, the majority of current breast cancer diagnostic research is devoted to determining whether a mammogram is malignant or benign. However, those frameworks did not consider how hidden layers affect the fusion layer before, after, or between them, which can be inaccurate and lead to a poor learning process. Thus, learning two examined-auxiliary (EA) low-dimensional and high-dimensional spaces can more effectively exploit the complex correlations between EA image pairings, producing higher-quality reconstruction results.

In summary, the main contributions of our work are as follows:
\begin{enumerate}
\item We introduce various mutations of multi-view networks to understand how the effectiveness of fusion at multiple positions in architecture. We proposed five strategies for fusion types: Pre Fusion, Early Fusion, Middle Fusion, Last Fusion, and Post Fusion. Additionally, we use ResNet-18 in our instance as a backbone and then divide it into different strategies to assess its efficacy.

\item We proposed a robust fusion block using the two well-known functions: average and concatenate as the aggregation operations and ablate the effect of skip connection between different views and fused features.

\end{enumerate}

\section{Methodology}

\subsection{Fusion Type}

Previously, many mammogram classification approaches have simply let the images go through the common-shared feature extractors to learn their information. Then, they continuously extend their methodology to increase performance. Afterward, they fused those features and fed them into various multiple perceptron layers (MLP) at last to produce the prediction. However, this can cause poor learning processing. Because of two different views of information, fused features can have specific noise on each view when combined together. This motivates us to propose new strategies of multi-view networks that empirically extracted before, after, or between fused features. Those approaches separately cut down the original backbone architecture into two parts: Coarse Layer (low-level individual extractor) and Fine Layer (high-level fusion extractor), which are the layers before the fusion block and the layers after the fusion block, respectively. 

\begin{figure*}[t]
	\centering
	\includegraphics[width=0.8\linewidth]{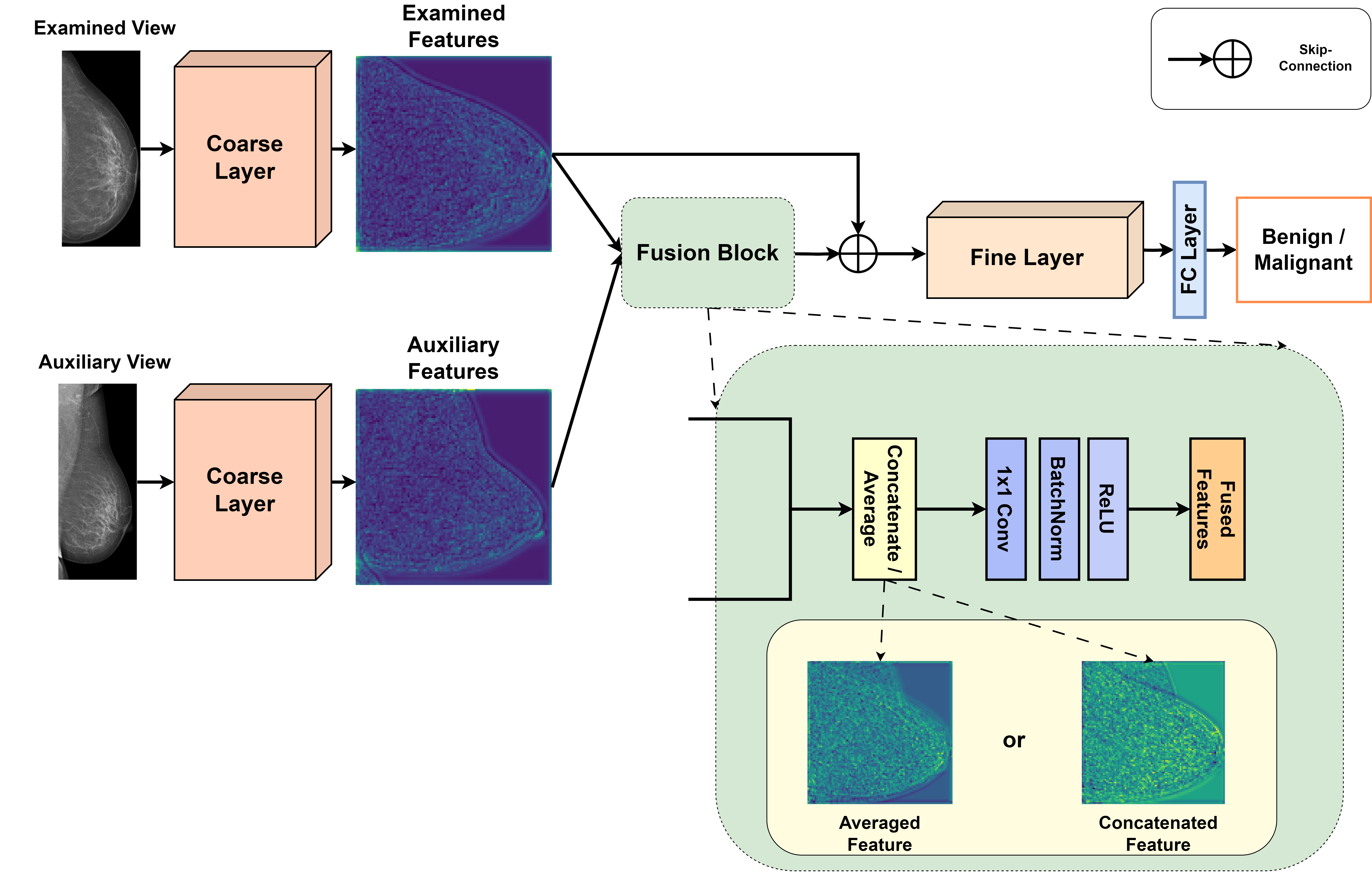}
	\caption{The overview of Ipsilateral Multi-View Network architecture. The two examine and auxiliary views are fed into the Coarse Layer, then fused together in the Fusion Block. Afterward, it goes through the Fine Layer and finally classification in the FC Layer.}
	\label{fig: framework1}
\end{figure*}

To be used as a building block of our network, ResNet-18 \cite{resnet} was used to be the main architecture, which consists of a pre-trained convolutional neural network. According to \emph{He et al.(2016)} \cite{resnet}, ResNet was originally built with many blocks in many different channel dimension sizes, which enable it to split apart. As shown in \emph{Fig.~\ref{fig: framework1}}, the input mammograms are ﬁrst fed into the Coarse Layer. Following that, the fusion block, average or concatenate, is subsequently applied to make the unchanging flow and combine in the backbone. The combined feature is then passed to the Fine Layer to extract the high-level features for distinguishing among the classes at the last classification layer. Therefore, there are five positions to separate and two aggregation functions which a total of ten models are discussed and evaluated in \emph{Section.~\ref{Results}}.

\begin{figure*}[ht]
	\centering
	\includegraphics[width=1\linewidth]{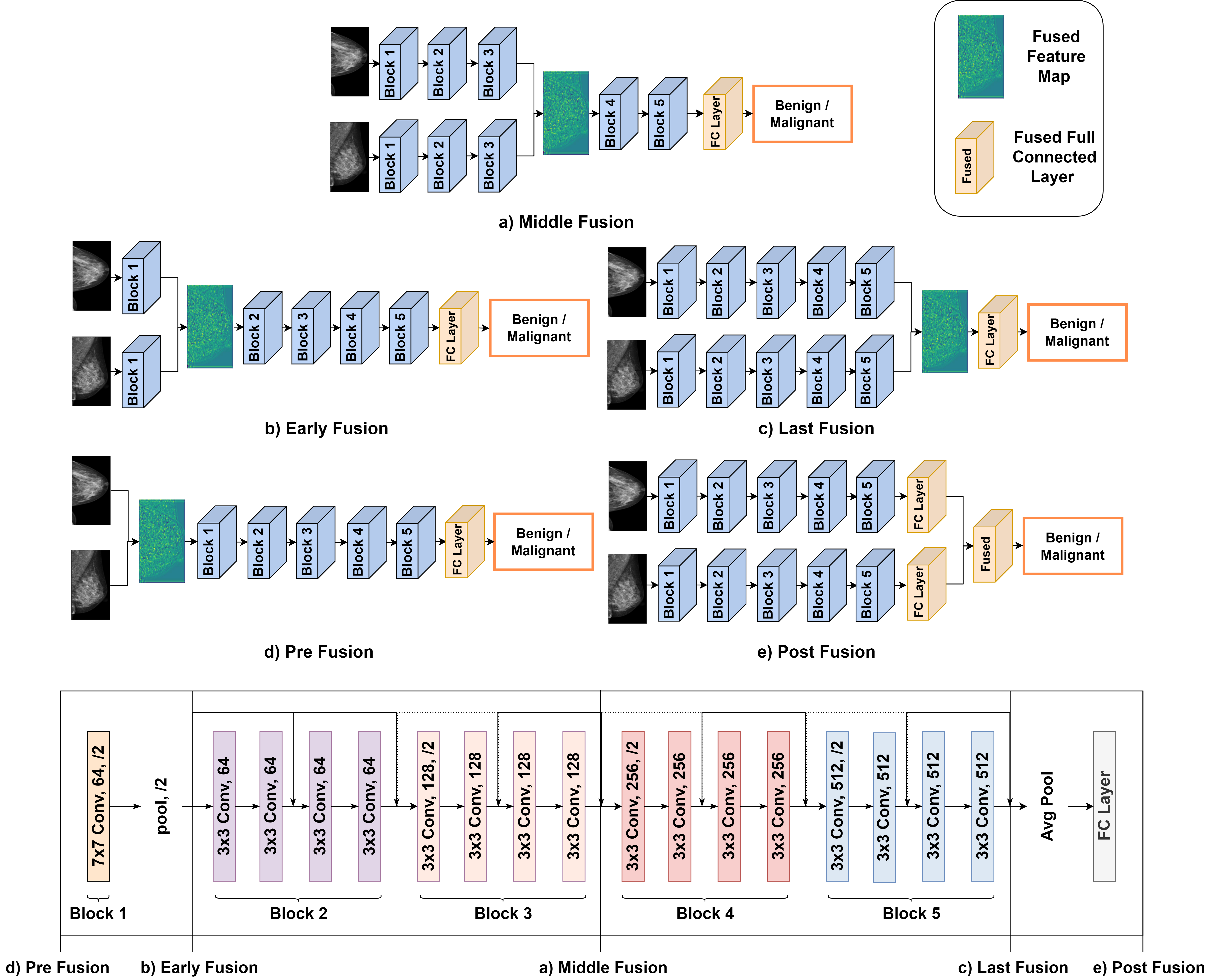}
	\caption{Five proposed fusion types (top) and ResNet-18 architecture with the fusion separated line (bottom). (a) Middle Fusion separates between block 3 and block 4 which contributes equally in both layers.  (b) Early Fusion separates between block 1 and block 2 of the backbone. (c) Last Fusion, in contrast with Early Fusion, separates most of the blocks in the backbone to the Coarse Layer, which cuts between block 5 and the average pooling function. (d) Pre Fusion put all layers of the backbone into the Fine Layer, nothing for the Coarse Layer.  (e) Post Fusion, in contrast with Pre Fusion, put all layers of the backbone into the Coarse Layer, nothing for the Fine Layer.}
	\label{fig: fused}
 \vskip -5pt
\end{figure*}

\subsubsection{Pre - Fusion (PreF)} \emph{(Fig \ref{fig: fused}.d)}
% (\emph{Fig.~\ref{fig: framework}a.})
This is the simplest and fewest computational resources model. Pre Fusion works as two screening mammography views individually not going through any layers. Instead, they are firstly aggregated together at the beginning without learning the mapping from low-dimensional and then passed the combined feature to the fully Fine Layer. This approach focuses mostly on the Fine Layer and removes totally process in Coarse Layer.
\subsubsection{Early - Fusion (EF)} \emph{(Fig \ref{fig: fused}.b)}
The separation now shifts the position to the right a little bit compared to PreF and locates between block 1 and block 2 in the feature extractor. The two views input now can be learned individually from all layers in block 1 in the Coarse Layer. This enhances the combined feature to perform better results when going through the Fine Layer. Fine Layer consisting of block 2, block 3, block 4, and block 5 of the backbone.
\subsubsection{Middle - Fusion (MF)} \emph{(Fig \ref{fig: fused}.a)}
In order to improve the performance and make full advantage of deep learning technologies to automatically boost resolution. Middle Fusion is shifted between block 3 and block 4 which have the same number of layers in Fine Layer and Coarse Layer. This can be called the most balanced type in various proposed strategies, which let features learn equally in both Layers.  
\subsubsection{Last - Fusion (LF)} \emph{(Fig \ref{fig: fused}.c)}
In LF, the aggregation locates after block 5 and before the average pooling layer. This is the final type we use 1x1 convolution to make an equal number of channel dimensions in two aggregation types which will be discussed later in \emph{Section.~\ref{fusionblock}}.
\subsubsection{Post - Fusion (PostF)} \emph{(Fig \ref{fig: fused}.e)}
After going through all layers individually in ResNet-18, we fuse two features of two views. This strategy learns a lot of individual information about two views. However, afterward, they do not have learnable parameters behind to extract the high-dimensional information. Post Fusion, in this case, is most likely with many traditional feature extractors that are used regularly. We modified a little bit compared with the traditional backbone to satisfy with various approaches above.

\subsection{Fusion Block} \label{fusionblock}
% \begin{figure*}[t]
% 	\centering
% 	\includegraphics[width=1\linewidth]{figures/Framework_main2_2.png}
% 	\caption{The average fusion block (left) fuses two ipsilateral views features using the average function. The concatenate fusion block (right), in contrast, fused two CC and MLO features using the concatenation function.}
% 	\label{fig: framework2}
% \end{figure*}

There are two main aggregation types that are mainly used in several methods nowadays: concatenate and average. Because two individual ipsilateral views are learned from two shared weight Fine Layer. Therefore, they need the combination function to continuously adapt the remained Coarse Layer. In the average fusion block, the average function uses the pixel-wise additional and then pixel-wise divided by two to perform the combined feature. In the concatenate fusion block, the concatenation function combines two features in the Coarse Layer along the channel dimension if there are feature maps that occur in PreF, EF, MF, and LF approaches. If there are flattened features in the PostF strategy, the fused feature will be obtained along the node size in a dense layer. These functions can be formulated in the following form:

\begin{equation}
    a_{avg-fused} = \frac{\emph{C}(x_{CC}) + \emph{C}(x_{MLO})}{2}
\end{equation}

\begin{equation}
    a_{concat-fused} = [\emph{C}(x_{CC}), \emph{C}(x_{MLO})] 
\end{equation}

where \emph{C(.)} is the Coarse Layer network, $a_{avg-fused}$,  $a_{concat-fused}$ is the averaged feature and concatenated feature in the fusion block, respectively, and $[.]$  is the concatenation function.

To alleviate the inconsistency in our model, the 1x1 convolution layer was added to keep the stabilization of the feature extractor in depth dimension flow. Because of the double number of channels in concatenation, we need 1x1 convolution to resize its channel dimension by half and continuously go through the remaining layers. In average block fusion, we do not need halving because it used pixel-wise calculation. Therefore, we keep the same dimensional in the 1x1 convolution layer to stabilize the flow. Furthermore, we used batch normalization \cite{batchnorm} and ReLU activation layers \cite{relu} in the fusion block after 1x1 conv2d was used for normalizing the fused features.  

\subsection{Class Imbalanced}

Dealing with unbalanced data is one of the biggest difficulties in medical image analysis. The issue is considerably more obvious in the area of medical imaging. 
% An unstable segmentation network that is skewed toward the class with a large area might result from segmentation training on data that is unbalanced by class. As a result, the selection of loss functions in segmentation networks is critical, especially when dealing with extremely imbalanced situations. This issue may be resolved by changing the distributions of the training data.
We present the Focal Loss \cite{focal}, which pushes the model to down-weight simple instances in order to learn about hard samples and is defined by adding a modulating component to the cross-entropy loss and a class balancing parameter

\begin{equation}
FL(p_t) = -\alpha_t (1 - p_t)^\gamma \log(p_t) ,
\end{equation}

where $p_t$ is the predicted probability of the true class, $\alpha_t$ is the class balancing factor for the true class, and $\gamma$ is the focusing parameter that controls the degree of down-weighting for well-classified examples. 
% In the VinDr-Mammo dataset, we set $\alpha_{benign} = 1$ and  $\alpha_{malignant} = 4$ to force the model dramatically focus on Malignancy samples 4 times more than Benign samples via the backpropagation of the weights in the model.  

\section{Experimental Settings} \label{Experimental results}

\subsection{Dataset preparation}\label{AA}

\textbf{VinDr-Mammo:}\cite{vindr} is a large-scale full-field digital mammography dataset with 20000 images from 5000 patients and 4 views per patient. From 2018 to 2020, the dataset was collected and labeled by three radiologists with an average of 19 years of clinical experience.
Each image was annotated using a BI-RADS scale ranging from 1 to 5. Due to the heavy imbalance from BI-RADS 1 and inconsistent annotation from BI-RADS 3, a subset of VinDr-Mammo with BI-RADS 2, 4, and 5 was brought for assessment. We divided them into two classes: benign is a set of BI-RADS 2 samples, and suspicious for malignancy is a set of BI-RADS 4 and 5 samples. According to the original train-test splitting information from a metadata file, we used 4532 images for training and validating and 1132 images for testing, with 4676 benign and 988 suspicious samples. For multi-view training settings, each patient from two sets remains four views, with two views of each breast. The training stage and Inference stage require two views each for performing classification.

\textbf{The Chinese Mammography Database (CMMD):}\cite{cmmd} includes 5.202 screening mammogram images from 1.775 studies. We trained on 1.172 non-malignant mammograms and 2.728 malignant screening images with 85\%:15\% ratio splitting on the training set and test set. Furthermore, we employ stratified sampling, resulting in 498 benign and 1157 malignancy ipsilateral view samples on the training set and 88 benign and 205 malignancy ipsilateral view samples on the testing set.
\subsection{Detailed Training \& Evaluation Metrics}

In our settings, we used the same architecture (ResNet-18 \cite{resnet}) for the feature extractor part of the framework. In the data loading part, the images are loaded with a batch size of 32 (two views for each breast with a total of 16 breasts on one side). The model was trained for 200 epochs using SGD optimizer \cite{sgd} with an initial learning rate $1 \times 10^{−3}$ and decays by $0.1$ after $20, 40, 60$, and $80$ epochs. Our images are preprocessed by resizing them into 800 for both training and testing. In our case, macro F1 - Score is the appropriate evaluation metric. Furthermore, the area under the ROC Curve (AUC  ROC) also is used for measuring the models' performance under slightly imbalanced dataset training.

\section{Results and Ablation Studies} \label{Results}

\subsection{Ipsilateral Multi-View Network}

\begin{table}[ht]
\caption{Quantitative results (\%) using AUC-ROC and macro F1-Score with ResNet-18 on the VinDr-Mammo and CMMD datasets.}
\label{tab: fusion}
\vskip 0.1in
\begin{center}
\begin{tabular}{l@{\hskip 0.1in}l@{\hskip 0.1in}@{\hskip 0.1in}c@{\hskip 0.1in}c@{\hskip 0.1in}@{\hskip 0.1in}c@{\hskip 0.1in}c@{\hskip 0.1in}}
\toprule
 &
 & \multicolumn{2}{c} {\textbf{Average}}  
 & \multicolumn{2}{c} {\textbf{Concatenate}}   \\
\toprule
\textbf{DataSet} & \textbf{Fusion Type}  &  F1-Score & AUC-ROC & F1-Score & AUC-ROC \\
\midrule
\multirow{5}{*}{VinDr-Mammo} &    PreF        & 68.71    & 68.49   & 73.28    & 73.46     \\
&    EF          & 71.49    & 71.92   & 69.44	& 68.67 	 \\ 
&    MF          & 74.00    & 72.15   & \textbf{75.34} & 74.24\\ 
&    LF          & \textbf{74.09} & 74.47	& 74.91  & 74.61  \\ 
&    PostF       & 74.08    & 75.35	 & 74.48    & 73.23     \\ 
\midrule
\multirow{5}{*}{CMMD} &    PreF        & 78.45    & 81.82   & 75.74    & 76.92     \\
&    EF          & 79.92    & 80.02   & 72.23	& 76.64 	 \\ 
&    MF          & \textbf{81.45}    & 84.16   & \textbf{77.77} & 80.42\\ 
&    LF          & 80.67  & 82.52	& 77.02  & 79.92  \\ 
&    PostF       & 81.32    & 83.70	 & 76.98    & 79.12     \\ 
\bottomrule
\end{tabular}
\end{center}
\vskip -0.2in
\end{table}

Table \ref{tab: fusion} illustrates the experimental results of the five proposed methods with two fusion block types: average and concatenate on VinDr-Mammo and CMMD datasets. 
% We observed that the performance of the proposed concatenate fusion block had surpassed the average fusion block in terms of Macro F1-Score and slightly in ROC AUC. We conjecture that the main issue, in this case, is the way we combined the features. In the average fusion block, the pixel-wise average function alleviates the important information and then adds other information from the opposite view which can be acted as adding noise. Typically, it can lead to collapsed representations. In the concatenate fusion block, we only append with an opposite feature which perfectly keeps the information remaining and stabilizing in the Fine Layer before moving to prediction.
In the comparison in fusion type, it is also shown that the Middle Fusion achieves high results on the whole experiments. It shows a significant improvement on macro F1-Score by $(+5.29\%)$ on average and $(+2.06\%)$ on concatenate in VinDr-Mammo compared with the conventional technique, Pre Fusion. In CMMD, it also improves with around $(+3\%)$ on average and $(+2.03\%)$ on concatenate. In particular, Middle Fusion outperforms both fusion block types with two datasets, but the average Last Fusion achieves a little improvement on VinDr-Mammo with a lower only $0.09\%$ than Middle Fusion. This loosening can be ignored because the average Middle Fusion also gets a good result. As a result, Middle Fusion can produce better performance in both fusion block types. We conjecture that claim is due to the equity in layers flow in the Middle Fusion. Since the Coarse Layer and the Fine Layer contains the balance separation in the feature extractor ResNet-18, the low-level features can adequately learn individual two examined-auxiliary (EA) images before fusing them together. Later, the fused feature might include noise inside needed the completely Fine Layer with various layers to procedure them into useful knowledge before the classifier layers. 

% \begin{figure*}[t]
% 	\centering
% 	\includegraphics[width=1\linewidth]{figures/Fig_CoarseToFine.png}
% 	\caption{The average fusion block (left) fuses two ipsilateral views features using the average function. The concatenate fusion block (right), in contrast, fused two CC and MLO features using the concatenation function.}
% 	\label{fig: framework3}
% \end{figure*}

\subsection{Skip Connection}

Table \ref{tab: skip} shows the ablation studies on the skip connection in various ways: examined, auxiliary, and examined-auxiliary (EA) in two backbones: ResNet-18 and RestNet-34. The skip connection with examined sample outperformed all of those strategies which rounded (2.12\%-1.35\%) on VinDr-Mammo and (1.91\%-1.39\%) on CMMD compared to the baseline, no skip connection. The skip connection with two examined-auxiliary views resulted in poor performance, even with baseline no skip connection. 

\begin{table}[h]
\caption{Ablation studies of skip connection on different views with ResNet-18 and ResNet-34 on VinDr-Mammo and CMMD datasets}
\label{tab: skip}
\begin{center}
\begin{tabular}{c@{\hskip 0.1in}c@{\hskip 0.1in}c@{\hskip 0.1in}|c@{\hskip 0.1in}c@{\hskip 0.1in}|c@{\hskip 0.1in}c@{\hskip 0.1in}}
\toprule
 \multicolumn{3}{c} {\textbf{VinDr-Mammo}}  & \multicolumn{2}{c} {\makecell{\textbf{ResNet-18} \\ \textbf{MF}}} 
 & \multicolumn{2}{c} {\makecell{\textbf{ResNet-34} \\ \textbf{MF}}}     \\
\toprule
 Concatenate & \makecell{Skip Connection \\ (Examined)}   & \makecell{Skip Connection \\ (Auxiliary)} & \makecell{Macro \\ F1-Score} & \makecell{AUC \\ ROC} & \makecell{Macro \\ F1-Score} & \makecell{AUC \\ ROC}\\
\midrule
% \midrule
% Baseline & \checkmark&  &  &  &  75.98\\
\midrule
    \checkmark &   &  &  73.22 & 70.66 & 74.63 & 72.18\\ 
\cmidrule(){1-3} \cmidrule(lr){4-7}
    \checkmark & \checkmark    &  & \textbf{75.34} & 74.24 & \textbf{75.98} & 74.86\\ 
\cmidrule(){1-3} \cmidrule(lr){4-7}
    \checkmark & \checkmark    & \checkmark &  71.68 & 72.66 & 73.27 & 68.92\\ 
\midrule
\midrule
\multicolumn{3}{c} {\textbf{CMMD}}  & \multicolumn{2}{c} {\makecell{\textbf{ResNet-18} \\ \textbf{MF}}} 
 & \multicolumn{2}{c} {\makecell{\textbf{ResNet-34} \\ \textbf{MF}}}     \\
 \midrule
  Concatenate & \makecell{Skip Connection \\ (Examined)}   & \makecell{Skip Connection \\ (Auxiliary)} & \makecell{Macro \\ F1-Score} & \makecell{AUC \\ ROC} & \makecell{Macro \\ F1-Score} & \makecell{AUC \\ ROC}\\
  \midrule
\midrule
    \checkmark &   &  &  75.86 & 77.10 & 78.12 & 77.67\\ 
\cmidrule(){1-3} \cmidrule(lr){4-7}
    \checkmark & \checkmark    &  & \textbf{77.77} & 80.42 & \textbf{79.51} & 81.97\\ 
\cmidrule(){1-3} \cmidrule(lr){4-7}
    \checkmark & \checkmark    & \checkmark &  74.87 & 79.21 & 77.86 & 79.54\\ 
\bottomrule
\end{tabular}
\end{center}
\vskip -0.2in
\end{table}

\section{Conclusion}

In this paper, we delved into many variants of CNN-based multi-view networks for ipsilateral breast cancer analysis. We empirically validate various fusion strategies and two fusion blocks in the same configuration. The experimental results show that the Middle Fusion significantly outperformed the remaining methods on large-scale clinical datasets. The intuition is that Middle Fusion has balance layers in both Coarse Layer and Fine Layer which extract enough low-level individual dimension and high-level fusion dimension, respectively. In addition, concatenate fusion block also keeps the information in two EA features without alleviating its as the average function does. In future work, we plan to develop this simple methodology to address the issue of dependency on multi-view data, and partial multi-view learning via the optimizing domain in EA views.

\section{Acknowledgement}
This paper is partially supported by AI VIETNAM. We thank the lab from the Biomedical Engineering Department of National Cheng Kung University - Integrated MechanoBioSystems Lab (IMBSL) for providing the GPU to support the numerical calculations in this paper.

% Reference Section

\end{document}